\documentclass[pdflatex,sn-mathphys-num]{sn-jnl}%

\usepackage{graphicx}%
\usepackage{multirow}%
\usepackage{amsmath,amssymb,amsfonts}%
\usepackage{amsthm}%
\usepackage{mathrsfs}%
\usepackage[title]{appendix}%
\usepackage{xcolor}%
\usepackage{textcomp}%
\usepackage{manyfoot}%
\usepackage{booktabs}%
\usepackage{graphicx}   
\usepackage{cleveref}
\usepackage[utf8]{inputenc}

\usepackage{algorithmicx}%
\usepackage{algpseudocode}%
\usepackage{listings}%
\usepackage[ruled,vlined]{algorithm2e}
\usepackage{listings}%

\usepackage{subfig}
\usepackage{tikz}
\usetikzlibrary{shapes.geometric, arrows}

\tikzstyle{startstop} = [rectangle, rounded corners, minimum width=3cm, minimum height=1cm,text centered, draw=black, fill=red!30]
\tikzstyle{process} = [rectangle, minimum width=3cm, minimum height=1cm, text centered, draw=black, fill=blue!20]
\tikzstyle{decision} = [diamond, minimum width=3cm, minimum height=1cm, text centered, draw=black, fill=green!30]
\tikzstyle{arrow} = [thick,->,>=stealth]

\UseRawInputEncoding

\begin{document}

\title{Brain Tumor Classification from 3D MRI Using Persistent Homology and Betti Features: A Topological Data Analysis Approach on BraTS 2020}

\author*[1]{\fnm{Faisal Ahmed} }\email{ahmedf9@erau.edu}

\affil*[1]{\orgdiv{Department of Data Science and Mathematics}, \orgname{Embry-Riddle Aeronautical University}, \orgaddress{\street{3700 Willow Creek Rd}, \city{Prescott}, \postcode{86301}, \state{Arizona}, \country{USA}}}

\abstract{
Accurate and interpretable brain tumor classification from medical imaging remains a challenging problem due to the high dimensionality and complex structural patterns present in magnetic resonance imaging (MRI). In this study, we propose a topology-driven framework for brain tumor classification based on Topological Data Analysis (TDA) applied directly to three-dimensional (3D) MRI volumes. Specifically, we analyze 3D Fluid Attenuated Inversion Recovery (FLAIR) images from the BraTS 2020 dataset and extract interpretable topological descriptors using persistent homology.

Persistent homology captures intrinsic geometric and structural characteristics of the data through Betti numbers, which describe connected components (Betti-0), loops (Betti-1), and voids (Betti-2). From the 3D MRI volumes, we derive a compact set of 100 topological features that summarize the underlying topology of brain tumor structures. These interpretable descriptors enable the representation of complex 3D tumor morphology while significantly reducing data dimensionality.

Unlike many deep learning approaches that require large-scale training data, complex architectures, or extensive data augmentation, the proposed framework relies on computationally efficient topological features extracted directly from the 3D images. The resulting features are used to train classical machine learning classifiers, including Random Forest and XGBoost, for binary classification of high-grade glioma (HGG) and low-grade glioma (LGG).

Experimental results on the BraTS 2020 dataset demonstrate that the proposed approach achieves competitive performance while maintaining strong interpretability and computational efficiency. In particular, the Random Forest classifier combined with selected Betti features achieves an accuracy of 89.19\%. These findings highlight the potential of persistent homology as an effective and lightweight alternative for analyzing complex 3D medical images and performing interpretable brain tumor classification.

}




\maketitle

\section{Introduction}\label{sec1}

Brain tumors are among the most severe neurological disorders, posing significant challenges for early diagnosis and treatment planning. Magnetic Resonance Imaging (MRI) is widely used in clinical practice for brain tumor detection due to its superior soft tissue contrast and ability to capture complex anatomical structures. In particular, multi-modal MRI datasets such as the Brain Tumor Segmentation (BraTS) challenge datasets have become important benchmarks for developing automated tumor analysis methods \cite{menze2014multimodal,bakas2018identifying}.

Traditional machine learning and deep learning approaches have shown promising results for brain tumor classification and segmentation tasks. Convolutional neural networks (CNNs) and other deep architectures have been widely used to learn hierarchical representations from MRI images \cite{litjens2017survey}. However, these approaches often require large amounts of labeled data, extensive computational resources, and complex model architectures. Furthermore, deep learning models frequently suffer from limited interpretability, which is a critical concern in medical decision-making scenarios.

Topological Data Analysis (TDA) has recently emerged as a powerful mathematical framework for capturing intrinsic geometric and structural characteristics of complex data \cite{carlsson2009topology}. Persistent homology, a central tool in TDA, provides a multi-scale representation of the topological structure of data by quantifying features such as connected components, loops, and voids. These structures are described by Betti numbers, commonly referred to as Betti-0, Betti-1, and Betti-2, respectively \cite{edelsbrunner2008persistent, ahmed20253d}. In medical imaging, persistent homology has demonstrated promising capabilities in characterizing structural patterns in biomedical images while providing interpretable descriptors \cite{byrne2016persistent, ahmed2023tofi, ahmed2023topological, ahmed2023topo, ahmed2025topo, ahmed2025topological, yadav2023histopathological}.

Unlike conventional feature extraction techniques, TDA-based features capture the underlying topology of data rather than relying solely on intensity-based or texture-based representations. This makes topological descriptors particularly suitable for analyzing complex anatomical structures such as brain tumors. Moreover, TDA provides compact and interpretable features that can significantly reduce computational complexity compared to high-dimensional image representations.

\begin{figure*}[t!]
    \centering
    \subfloat[\scriptsize 2D PCA projection of Betti-0 features extracted from BraTS2020 MRI volumes showing separation between LGG and HGG samples.\label{fig:pca-B0}]{
        \includegraphics[width=0.45\linewidth]{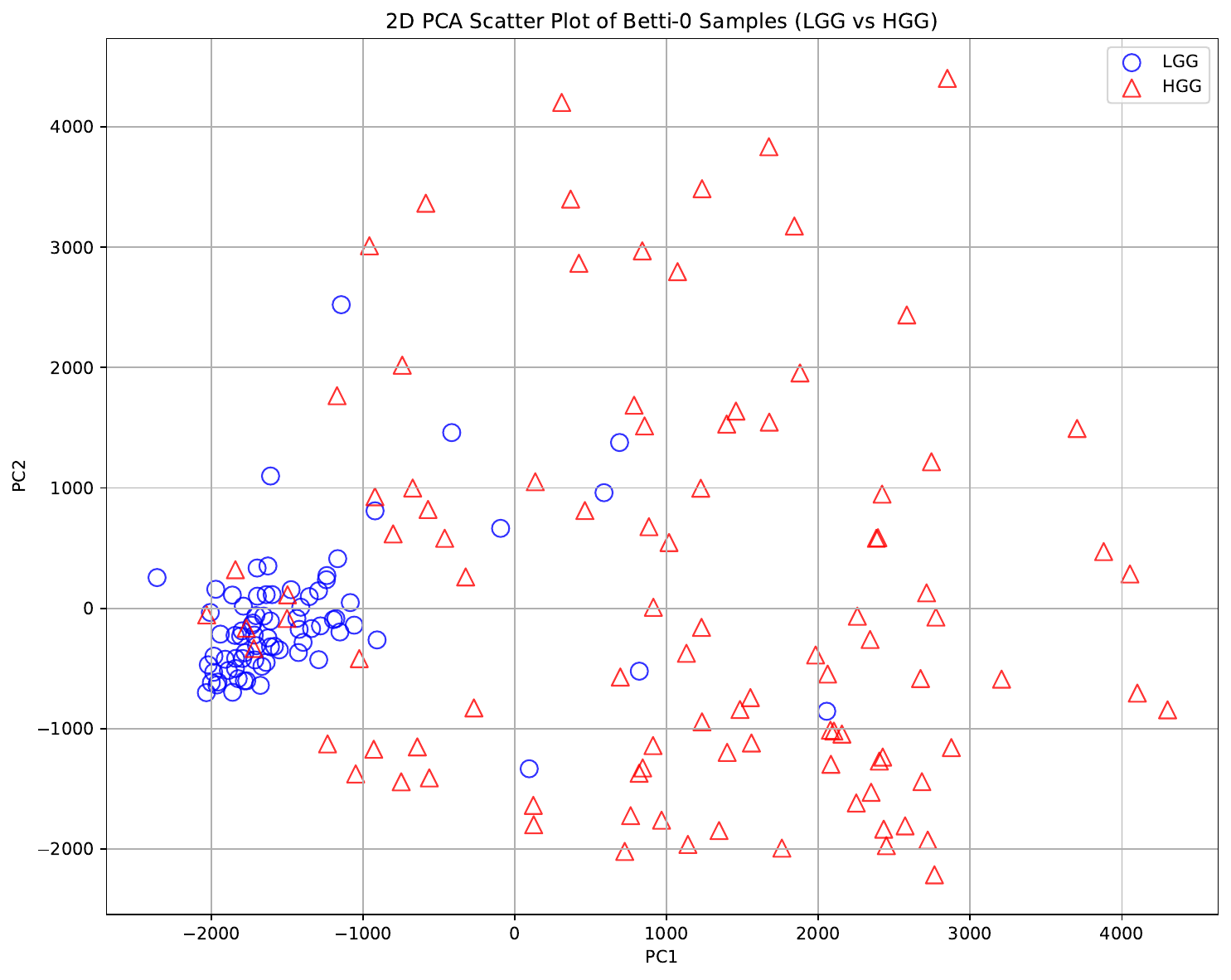}}
    \hfill
    \subfloat[\scriptsize 2D PCA projection of Betti-1 features highlighting structural differences between low-grade and high-grade gliomas.\label{fig:pca-B1}]{
        \includegraphics[width=0.45\linewidth]{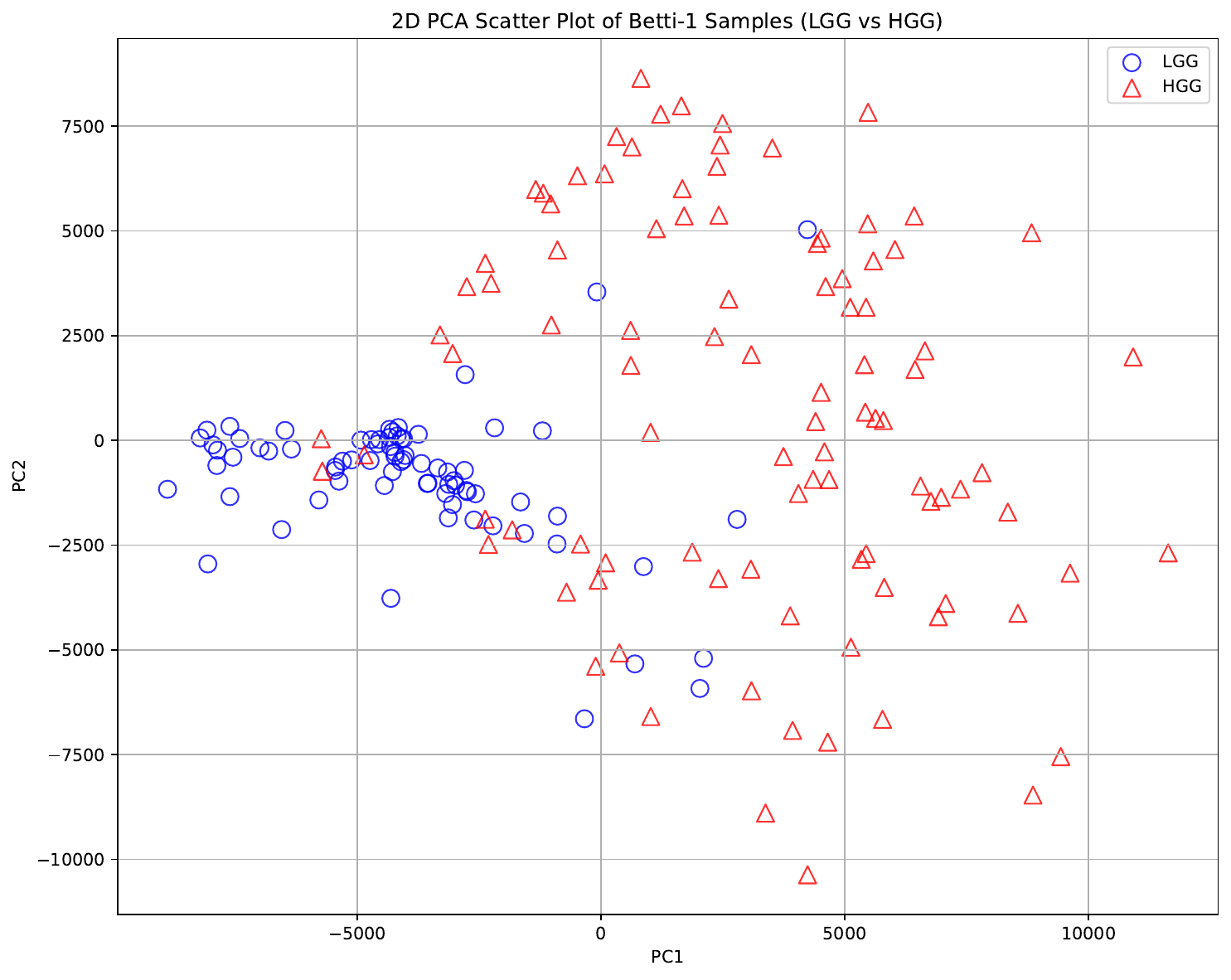}}
    
    \vspace{0.3cm}
    
    \subfloat[\scriptsize 2D PCA projection of Betti-2 features capturing cavity-related topological structures in 3D MRI volumes.\label{fig:pca-B2}]{
        \includegraphics[width=0.45\linewidth]{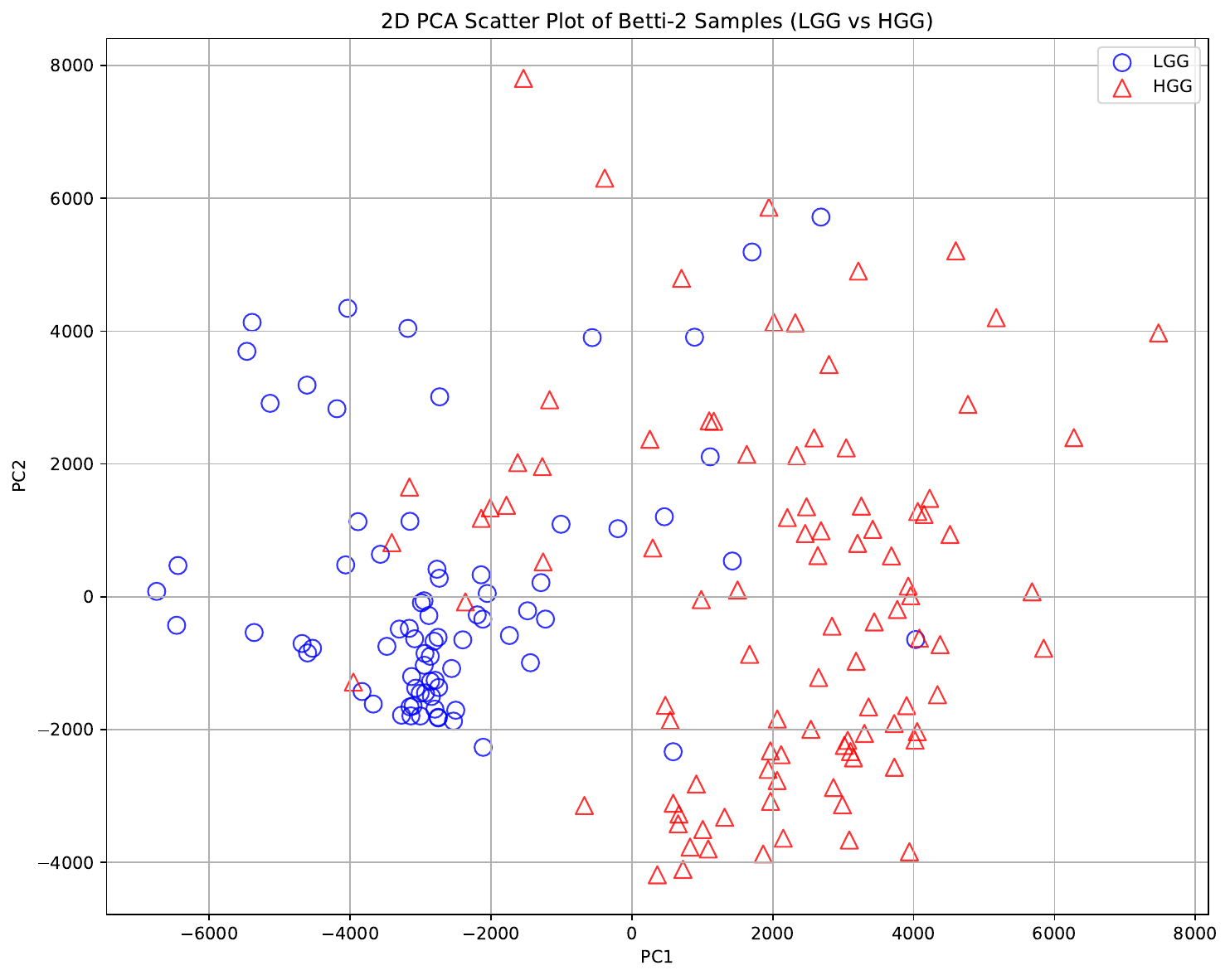}}
    
    \caption{\footnotesize Visualization of topological feature distributions using Principal Component Analysis (PCA). The plots show the first two principal components of Betti-based feature vectors derived from BraTS2020 MRI volumes: (a) Betti-0 features representing connected components, (b) Betti-1 features corresponding to loops or tunnels, and (c) Betti-2 features representing cavities. The separation between low-grade glioma (LGG) and high-grade glioma (HGG) samples demonstrates the discriminative capability of persistent homology-based topological descriptors for brain tumor classification.}
    
    \label{fig:pca-visualization}
\end{figure*}

\begin{figure*}[t!]
    \centering
    \subfloat[\scriptsize Median Betti-0 curves with $40\%$ confidence bands for LGG and HGG samples.\label{fig:B0-bands}]{
        \includegraphics[width=0.45\linewidth]{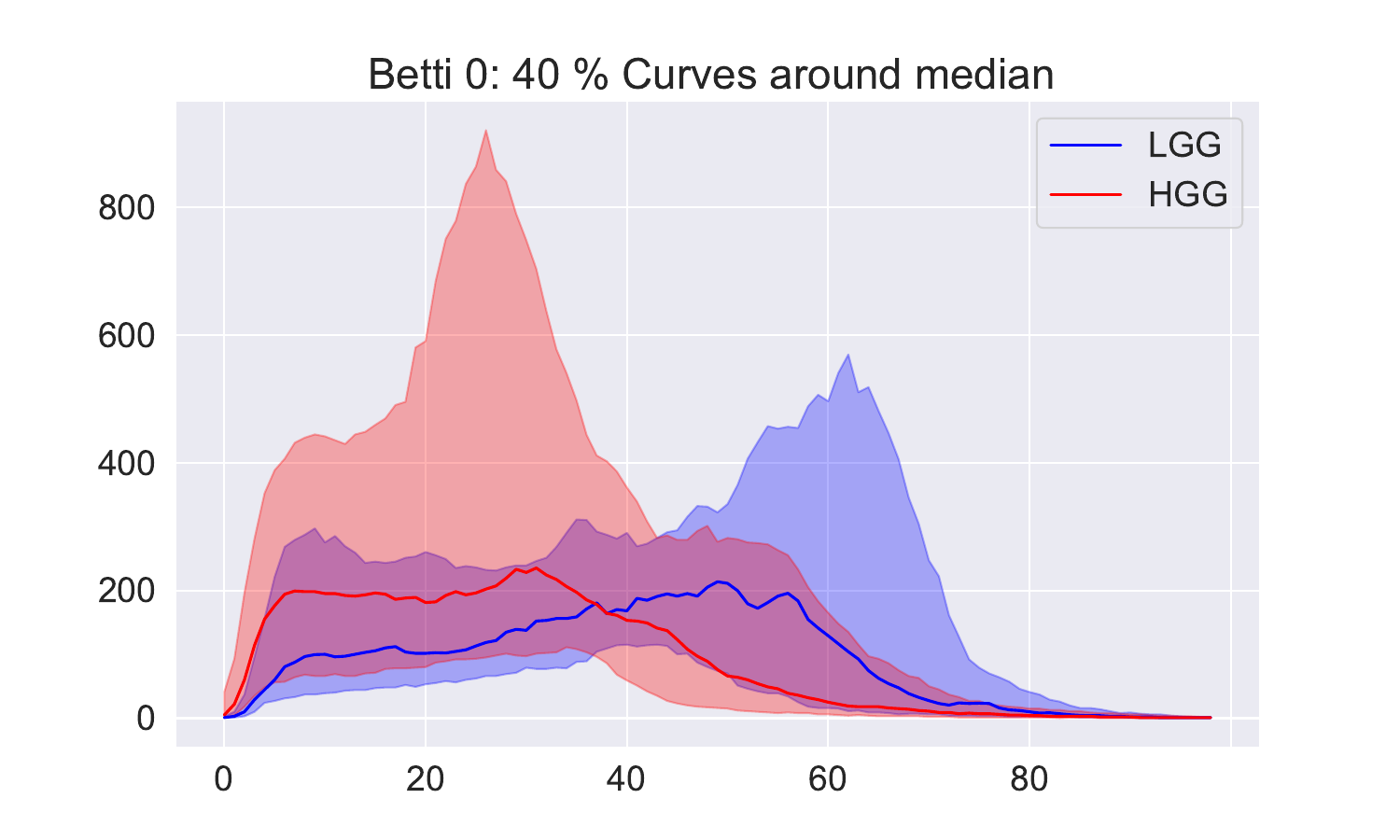}}
    \hfill
    \subfloat[\scriptsize Median Betti-1 curves with $40\%$ confidence bands for LGG and HGG samples.\label{fig:B1-bands}]{
        \includegraphics[width=0.45\linewidth]{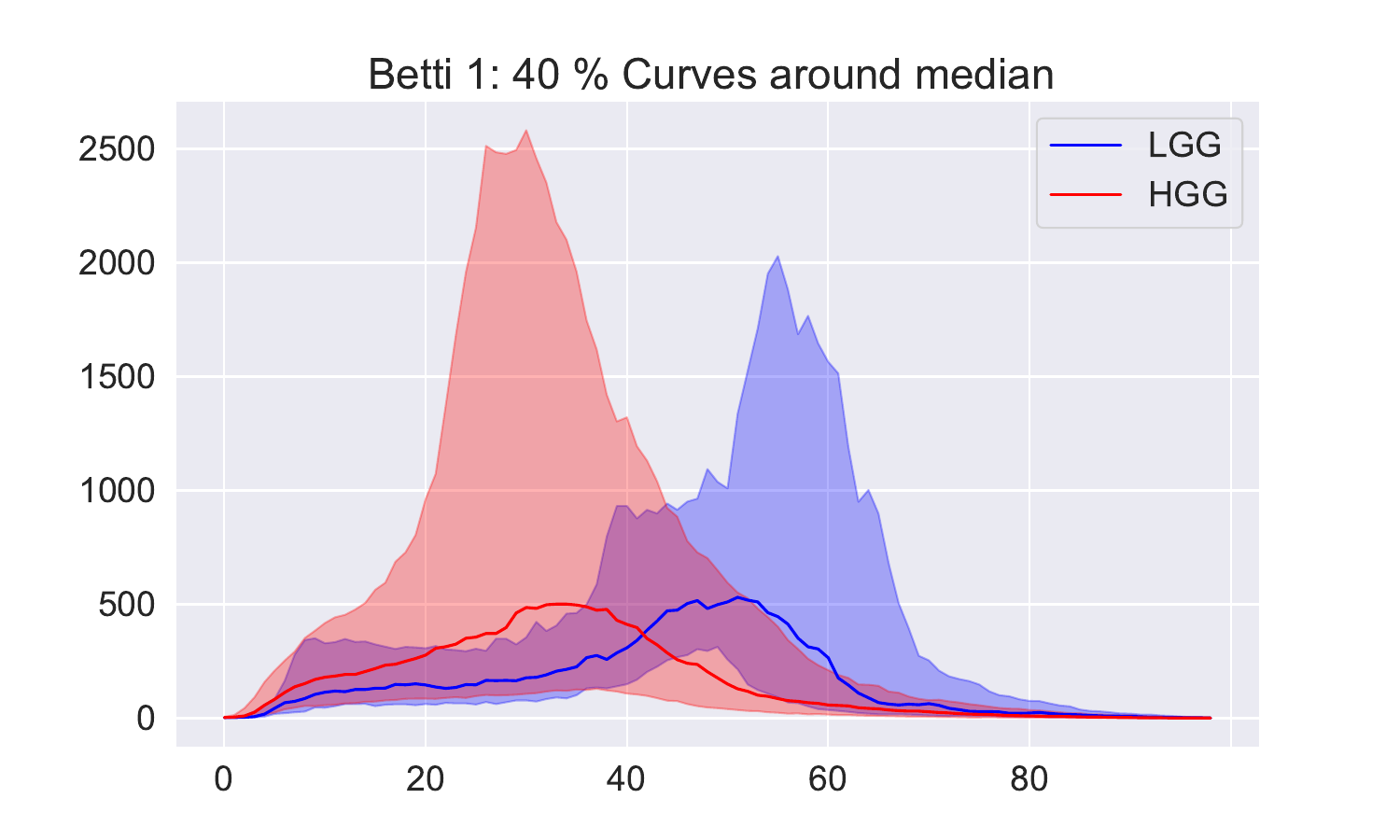}}
    
    \vspace{0.3cm}
    
    \subfloat[\scriptsize Median Betti-2 curves with $40\%$ confidence bands for LGG and HGG samples.\label{fig:B2-bands}]{
        \includegraphics[width=0.45\linewidth]{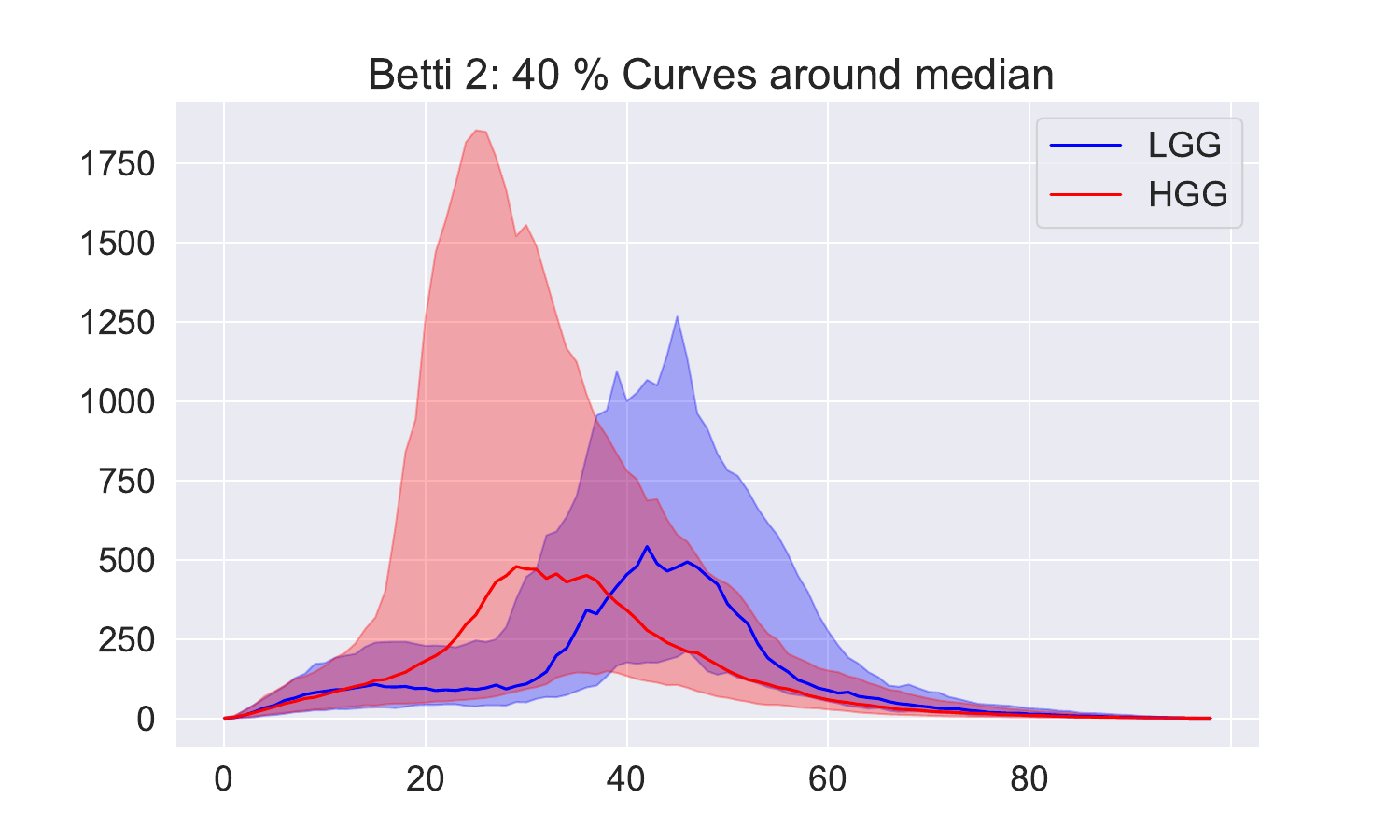}}
    
    \caption{\footnotesize Visualization of Betti functions extracted from 3D MRI volumes in the BraTS2020 dataset. Each plot shows the median Betti curve along with $40\%$ confidence bands for the two tumor classes: low-grade glioma (LGG) and high-grade glioma (HGG). The $x$-axis represents normalized grayscale filtration thresholds, while the $y$-axis represents the number of topological features at each threshold: connected components for Betti-0, loops for Betti-1, and cavities for Betti-2. These curves illustrate structural differences between LGG and HGG tumors captured by persistent homology.}
    
    \label{fig:Betti-bands}
\end{figure*}

\begin{figure*}[t!]
    \centering
    \subfloat[\scriptsize Violin plot of Betti-0 (connected components) feature distributions for LGG and HGG samples.\label{fig:vio-B0}]{
        \includegraphics[width=0.45\linewidth]{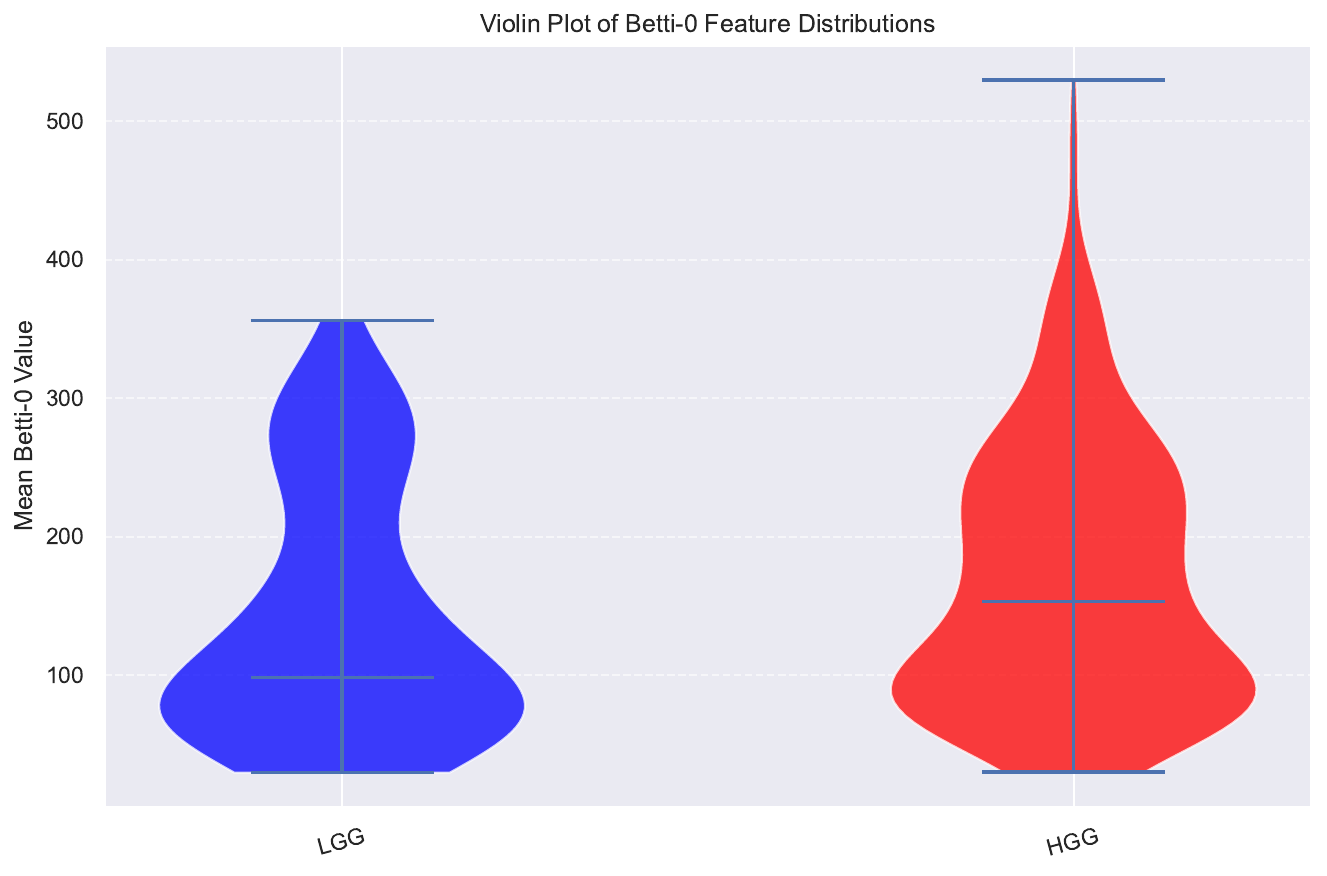}}
    \hfill
    \subfloat[\scriptsize Violin plot of Betti-1 (topological loops) feature distributions for LGG and HGG samples.\label{fig:vio-B1}]{
        \includegraphics[width=0.45\linewidth]{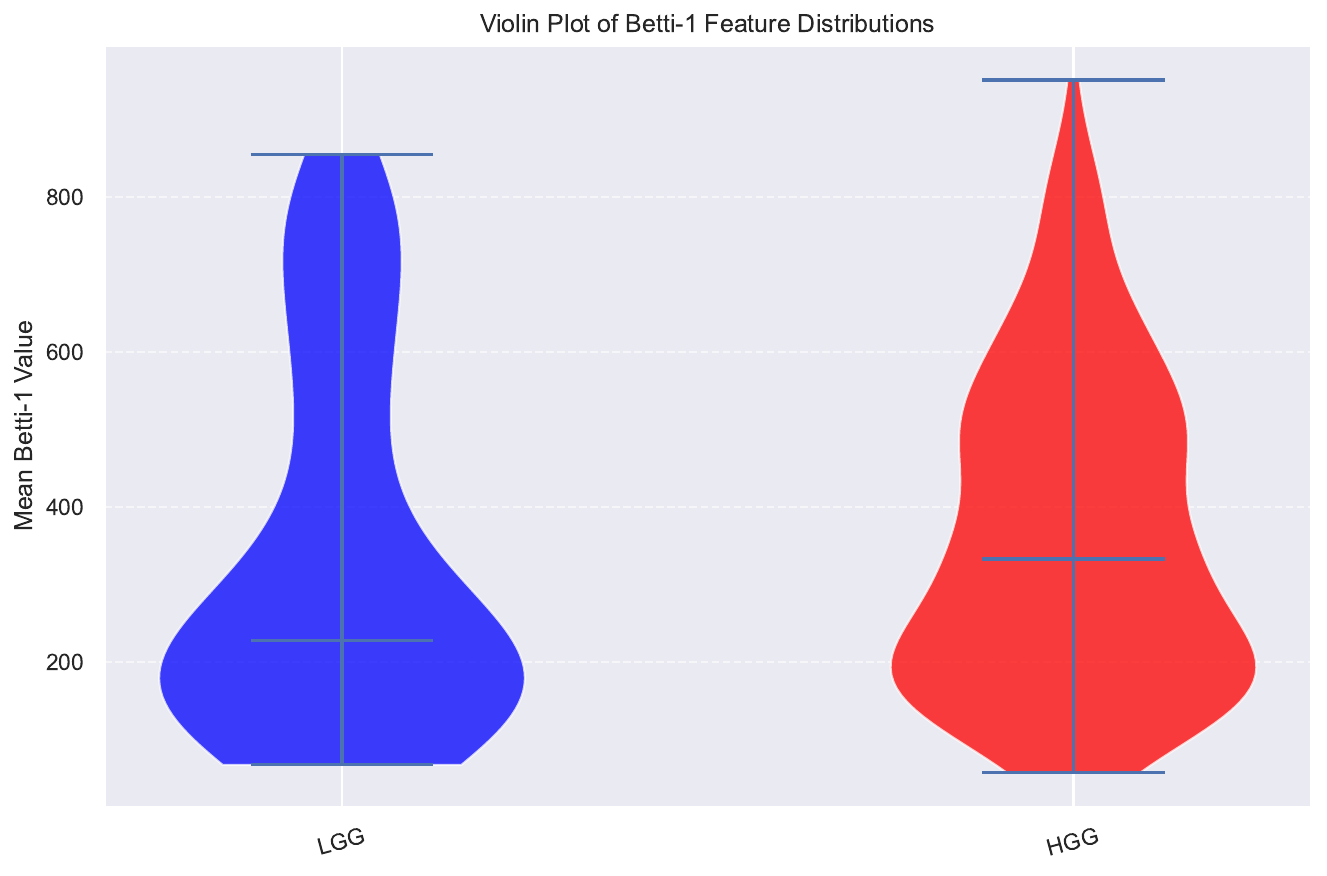}}
    
    \vspace{0.3cm}
    
    \subfloat[\scriptsize Violin plot of Betti-2 (topological cavities) feature distributions for LGG and HGG samples.\label{fig:vio-B2}]{
        \includegraphics[width=0.45\linewidth]{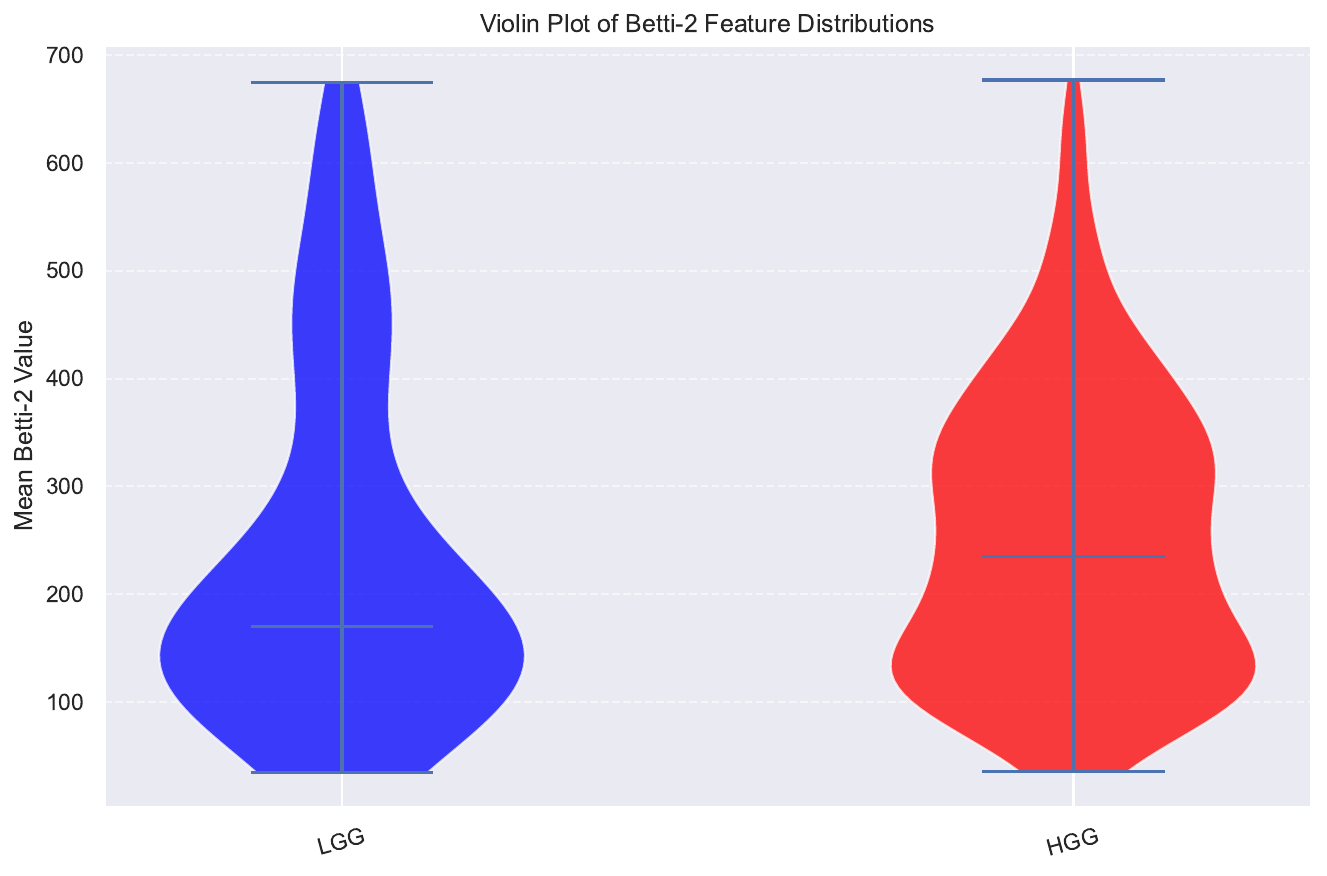}}
    
    \caption{\footnotesize Visualization of topological feature distributions derived from persistent homology applied to 3D MRI volumes in the BraTS2020 dataset. Each violin plot shows the distribution of aggregated Betti features for the two tumor classes: low-grade glioma (LGG) and high-grade glioma (HGG). (a) Betti-0 features represent connected components, (b) Betti-1 features correspond to loops or tunnels, and (c) Betti-2 features capture cavities within the MRI volume. Differences in the distributions between LGG and HGG indicate that topological descriptors extracted via persistent homology capture meaningful structural variations between tumor grades.}
    
    \label{fig:violin-visualization}
\end{figure*}

In this study, we propose a topology-driven framework for brain tumor classification using persistent homology applied directly to three-dimensional (3D) MRI volumes. Specifically, we utilize the FLAIR modality from the BraTS 2020 dataset and extract topological descriptors from full 3D images rather than two-dimensional slices. From these volumes, we compute Betti-based features (Betti-0, Betti-1, and Betti-2), which capture structural characteristics of tumor morphology in a compact and interpretable manner (\Cref{fig:pca-visualization}, \Cref{fig:Betti-bands}, \Cref{fig:violin-visualization}). These features are then used to train classical machine learning models for binary classification of high-grade glioma (HGG) and low-grade glioma (LGG).

\noindent\textbf{Our Contributions}
\begin{itemize}
    \item We propose a topology-driven framework for brain tumor classification using persistent homology applied directly to three-dimensional MRI volumes.
    
    \item We extract interpretable topological descriptors based on Betti-0, Betti-1, and Betti-2 features that capture structural characteristics of tumor morphology (\Cref{fig:pca-visualization}, \Cref{fig:Betti-bands}, \Cref{fig:violin-visualization}).
    
    \item The proposed method provides a computationally efficient alternative to deep learning approaches by using compact topological features without requiring data augmentation or complex model architectures.
    
    \item Extensive experiments on the BraTS 2020 dataset demonstrate that the proposed approach achieves competitive classification performance for distinguishing high-grade and low-grade gliomas.
\end{itemize}

\begin{figure}[t!]
\centering

\subfloat[\scriptsize Example LGG MRI slice from BraTS2020.\label{fig:lgg1}]{
\includegraphics[width=0.3\linewidth]{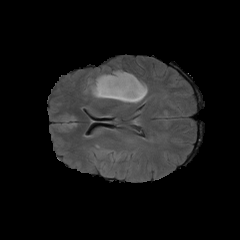}}
\hfill
\subfloat[\scriptsize Example LGG MRI slice from BraTS2020.\label{fig:lgg2}]{
\includegraphics[width=0.3\linewidth]{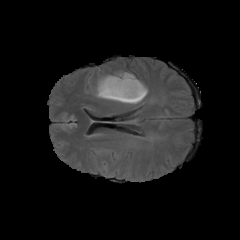}}
\hfill
\subfloat[\scriptsize Example LGG MRI slice from BraTS2020.\label{fig:lgg3}]{
\includegraphics[width=0.3\linewidth]{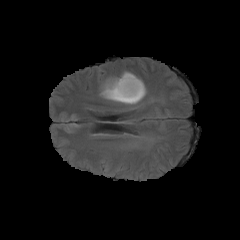}}

\vspace{0.2cm}

\subfloat[\scriptsize Example HGG MRI slice from BraTS2020.\label{fig:hgg1}]{
\includegraphics[width=0.3\linewidth]{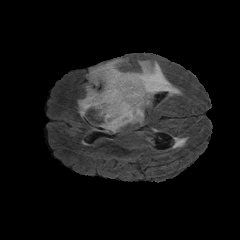}}
\hfill
\subfloat[\scriptsize Example HGG MRI slice from BraTS2020.\label{fig:hgg2}]{
\includegraphics[width=0.3\linewidth]{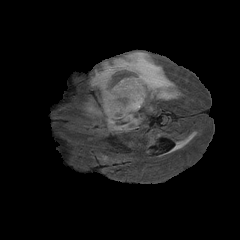}}
\hfill
\subfloat[\scriptsize Example HGG MRI slice from BraTS2020.\label{fig:hgg3}]{
\includegraphics[width=0.3\linewidth]{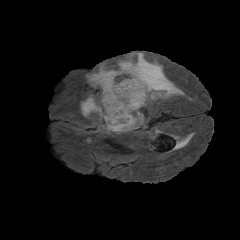}}

\caption{\footnotesize Representative 2D slices from 3D FLAIR MRI volumes in the BraTS2020 dataset. The first row shows examples of low-grade glioma (LGG) cases, while the second row shows examples of high-grade glioma (HGG) cases. These images illustrate the structural differences in tumor appearance between LGG and HGG samples used for topological feature extraction.}

\label{fig:image-samples}
\end{figure}

\section{Related Works}\label{sec2}

Automated analysis of brain magnetic resonance imaging (MRI) has been widely investigated for computer-aided diagnosis of neurological disorders and brain tumor assessment. The development of large public datasets such as the Brain Tumor Segmentation (BraTS) challenge has significantly accelerated research in automated brain tumor analysis by providing standardized multi-modal MRI data and evaluation protocols~\cite{menze2014multimodal,bakas2018identifying}. Early approaches primarily relied on handcrafted features derived from intensity statistics, texture descriptors, and morphological characteristics combined with conventional machine learning classifiers. Methods such as Support Vector Machines (SVMs) and Random Forests demonstrated promising performance in distinguishing tumor types by leveraging structural variations in MRI data~\cite{bauer2013survey}.

In recent years, deep learning has become the dominant paradigm for medical image analysis. Convolutional Neural Networks (CNNs) have been widely adopted for brain tumor detection, segmentation, and classification due to their ability to automatically learn hierarchical feature representations from imaging data~\cite{litjens2017survey}. Both two-dimensional (2D) and three-dimensional (3D) CNN architectures have been proposed for analyzing volumetric MRI data. In particular, 3D CNN models are capable of capturing spatial context across slices, which is important for modeling tumor morphology in brain MRI~\cite{cciccek20163d}. Despite their strong performance, deep learning models often require large annotated datasets, extensive data augmentation, and substantial computational resources, which can limit their practical applicability in clinical environments.

To address the limitations of purely data-driven approaches, researchers have explored alternative feature representations that capture structural characteristics of medical images in a more interpretable manner. Topological Data Analysis (TDA) has recently emerged as a powerful mathematical framework for analyzing complex high-dimensional data by studying their intrinsic geometric and topological structures~\cite{carlsson2009topology}. Persistent homology, one of the key tools in TDA, provides a multi-scale representation of the topological properties of data by tracking the evolution of connected components, loops, and voids across different scales~\cite{edelsbrunner2008persistent}. These topological structures are quantified using Betti numbers, which describe fundamental shape characteristics of the data.

TDA has been successfully applied to a variety of biomedical imaging tasks, including histopathology analysis, neuroimaging, and volumetric medical image analysis. Previous studies have demonstrated that topological descriptors derived from persistent homology can effectively capture global structural patterns that are difficult to represent using conventional intensity-based features~\cite{qaiser2019persistent, byrne2016persistent}. An increasing number of studies have highlighted the utility of topological data analysis (TDA) in medical imaging, with successful applications across brain MRI, histopathology, and volumetric imaging~\cite{ahmed2026four, ahmed2025topo, ahmed2023topo, ahmed2023topological, ahmed2023tofi, ahmed20253d, yadav2023histopathological, ahmed2025topological, ahmed2026hybrid}. In brain MRI analysis, topological features have shown promise in characterizing complex anatomical structures and providing robust representations that remain stable under noise and small perturbations. Recent studies have explored transfer learning and transformer-based architectures for medical image analysis, yet performance often degrades in small or imbalanced datasets~\cite{ahmed2025colormap, ahmed2025hog, ahmed2025ocuvit, ahmed2025robust, ahmed2025histovit, ahmed2025addressing, ahmed2025repvit, ahmed2025pseudocolorvit, rawat2025efficient, ahmed2025transfer}.

Compared with deep learning approaches, TDA-based methods offer several advantages, including improved interpretability, lower computational complexity, and reduced dependence on large-scale annotated datasets. These properties make topological descriptors particularly attractive for medical imaging tasks where data availability and model interpretability are critical considerations. Motivated by these advantages, this work explores the use of persistent homology to extract interpretable Betti-based features from three-dimensional (3D) MRI volumes for brain tumor classification. By leveraging the topological structure of the data, the proposed framework aims to provide an efficient and interpretable alternative to conventional deep learning approaches for analyzing complex brain tumor morphology.

\begin{figure}[t]
    \centering
    \includegraphics[width=\linewidth]{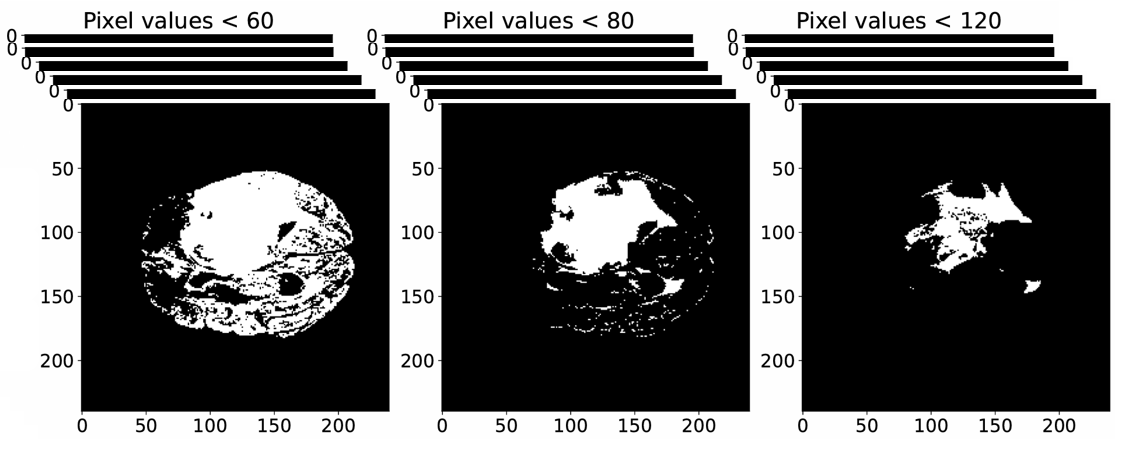}
    \caption{\small 
    \textbf{Cubical sublevel filtration of an MRI slice.} 
    A grayscale 3D MRI is progressively thresholded to generate a sequence of binary images. The images $\mathcal{X}_{60}$, $\mathcal{X}_{80}$, and $\mathcal{X}_{120}$ correspond to threshold values of $60$, $80$, and $120$, respectively. As the threshold increases, additional voxels become activated, producing a nested sequence of cubical complexes. During this filtration process, topological structures such as connected components ($k=0$), loops or tunnels ($k=1$), and cavities ($k=2$) appear and evolve, which are subsequently captured by persistent homology.
    }
    \label{fig:filtration}
\end{figure}

\section{Method}

In this study, we propose a topological framework for classifying brain tumors from the BraTS2020 dataset using three-dimensional (3D) MRI volumes (\Cref{fig:image-samples}). Each MRI volume has spatial dimensions $155 \times 240 \times 240$, corresponding to depth, height, and width respectively. Instead of directly training machine learning models on high-dimensional voxel intensities, we extract informative topological descriptors using Persistent Homology (PH), a key technique in Topological Data Analysis (TDA). These descriptors capture intrinsic geometric and structural patterns within MRI volumes across multiple spatial scales.

\begin{figure*}[t!]
    \centering
    \includegraphics[width=\linewidth]{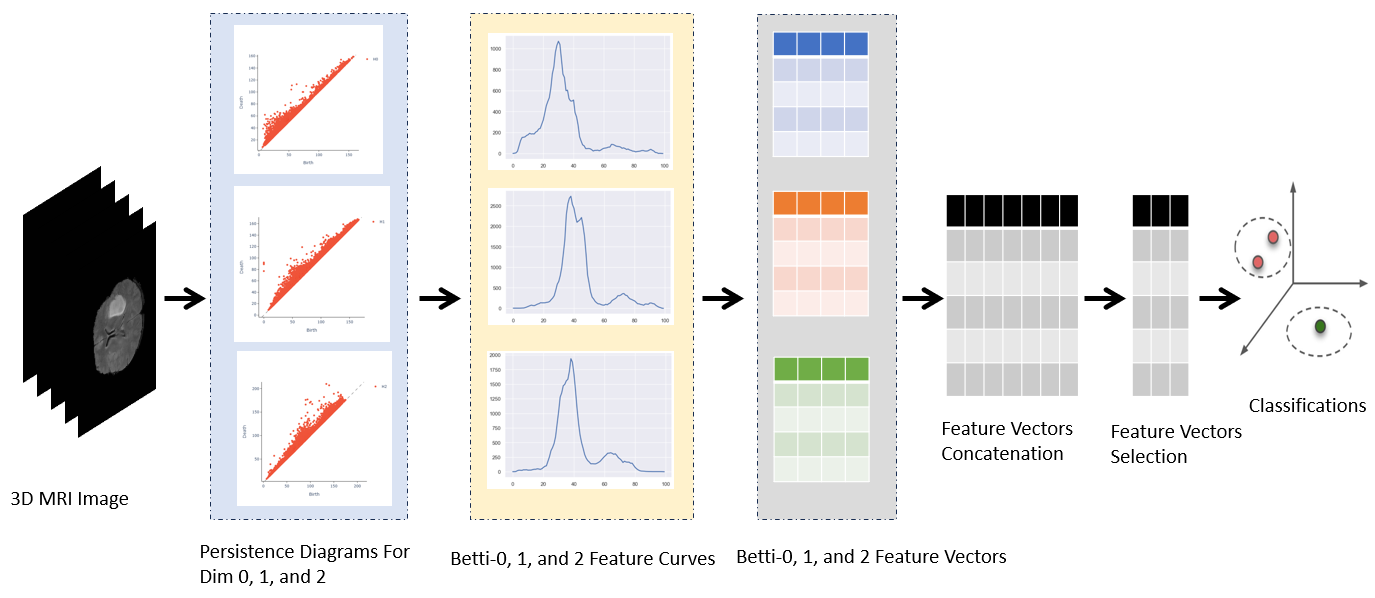}
    \caption{\footnotesize 
    \textbf{Overview of the proposed TDA-based classification framework.} 
    The pipeline begins with a 3D MRI volume composed of multiple 2D slices. Persistent homology is then applied using cubical filtration to compute persistence diagrams for homology dimensions $k=0,1,$ and $2$, capturing connected components, loops, and cavities within the image structure. These diagrams are subsequently transformed into Betti curves, which describe the evolution of topological features across the filtration scale. The resulting Betti-0, Betti-1, and Betti-2 curves are converted into fixed-length feature vectors and concatenated to form a unified topological representation for each MRI volume. Feature selection is then performed to identify the most informative descriptors, and the selected features are finally used as input to machine learning classifiers to distinguish between low-grade glioma (LGG) and high-grade glioma (HGG) in the BraTS2020 dataset.
    }
    \label{fig:flowchart}
\end{figure*}

The overall framework consists of four main stages: (i) construction of filtrations from MRI volumes, (ii) computation of persistence diagrams, (iii) vectorization of topological features using Betti curves, and (iv) classification using machine learning models with feature selection. An overview of the proposed pipeline is illustrated in \Cref{fig:flowchart}.

\subsection{Persistent Homology for 3D MRI Volumes}

Let $X \subset \mathbb{R}^3$ denote a 3D grayscale MRI volume consisting of voxels 
$\Delta_{ijk}$ with corresponding intensity values $\Upsilon_{ijk}$. Formally,

\[
X = \{\Delta_{ijk} \mid 1 \le i \le 155,\; 1 \le j \le 240,\; 1 \le k \le 240\}.
\]

Since adjacent MRI slices often contain redundant information, we focus on the central portion of each volume where tumor structures are most prominent. Specifically, from the total $155$ slices, we select slices indexed from $30$ to $90$, resulting in a reduced volume that preserves relevant anatomical structures while reducing computational complexity. Thus, the effective volume used for analysis has dimensions approximately $60 \times 240 \times 240$.

\subsection{Step 1: Constructing Filtrations}

Persistent homology analyzes the evolution of topological structures by constructing a filtration of cubical complexes from the grayscale MRI volume (\Cref{fig:filtration}). Let $t_1 < t_2 < \dots < t_N$ denote a sequence of intensity thresholds. Using these thresholds, we define a sublevel filtration

\[
X_1 \subset X_2 \subset \dots \subset X_N,
\]

where each cubical complex is defined as

\[
X_n = \{ \Delta_{ijk} \subset X \mid \Upsilon_{ijk} \le t_n \}.
\]

In this process, voxels are gradually activated as the threshold increases, producing a nested sequence of binary volumes. This sequence forms a filtration in which the topology of the image evolves across intensity levels. The emergence and disappearance of topological structures during this process capture important geometric characteristics of the MRI data. An illustration of this cubical sublevel filtration is shown in \Cref{fig:filtration}.

Although sublevel filtration is used in this work, superlevel filtration could also be applied by activating voxels in decreasing order of grayscale intensity. Both approaches encode equivalent topological information due to Alexander duality.

\subsection{Step 2: Persistence Diagrams}

Persistent homology tracks the birth and death of topological structures throughout the filtration process. For each dimension $k$, the persistence diagram is defined as the multiset

\[
PD_k(X) = \{(b_\sigma, d_\sigma) \mid \sigma \in H_k(X_n), \; b_\sigma \le n \le d_\sigma \},
\]

where $b_\sigma$ and $d_\sigma$ represent the birth and death filtration indices of a topological feature $\sigma$, and $H_k(X_n)$ denotes the $k$-th homology group of the cubical complex $X_n$.

For three-dimensional MRI data, the most relevant topological features arise in homology dimensions $k=0,1,$ and $2$. In particular, $k=0$ corresponds to connected components, which represent the number of distinct regions that appear during filtration. The dimension $k=1$ captures loops or tunnels within the image structure, while $k=2$ represents cavities or voids corresponding to enclosed volumetric regions. These features provide a multi-scale topological description of the anatomical structures present in the MRI volume.

The persistence of each topological feature is defined as

\[
\text{pers}(\sigma) = d_\sigma - b_\sigma.
\]

Features with larger persistence correspond to dominant structural patterns in the image, whereas features with shorter persistence often correspond to minor variations or noise.

\subsection{Step 3: Topological Feature Vectorization}

Persistence diagrams consist of sets of birth–death coordinate pairs and therefore cannot be directly used as input to machine learning models. To obtain fixed-length numerical representations, we transform persistence diagrams into Betti curves.

The $k$-th Betti number at threshold $t$ is defined as

\[
\beta_k(t) = \text{rank}(H_k(X_t)),
\]

which counts the number of $k$-dimensional topological features present at threshold $t$.

In this study, the grayscale interval $[0,255]$ is uniformly discretized into $N=100$ filtration thresholds. For each threshold $t_n$, the Betti numbers are computed to form Betti vectors

\[
\boldsymbol{\beta}_k(X) =
[\beta_k(t_1), \beta_k(t_2), \dots, \beta_k(t_{100})].
\]

Since Betti curves are computed for homology dimensions $k=0,1,$ and $2$, the final topological representation of each MRI volume is obtained by concatenating these vectors:

\[
\mathbf{f}(X) =
[\boldsymbol{\beta}_0(X), \boldsymbol{\beta}_1(X), \boldsymbol{\beta}_2(X)]
\in \mathbb{R}^{300}.
\]

Thus, each MRI volume is represented by a $300$-dimensional feature vector capturing multi-scale topological information describing connected components, loops, and cavities within the brain structure.

\subsection{Feature Selection}

Although the Betti vectors capture rich structural information, some features may be redundant or less informative for classification. To reduce dimensionality and improve model efficiency, we apply feature selection based on feature importance scores obtained from tree-based models.

Let

\[
\mathbf{F} = \{f_1,f_2,\dots,f_{300}\}
\]

denote the set of extracted Betti features. For a trained ensemble model, each feature $f_j$ is assigned an importance score $I_j$ representing its contribution to the classification task.

A subset of features is selected using a threshold parameter $\tau$ defined as

\[
S_\tau = \{f_j \in \mathbf{F} \mid I_j \ge \tau\}.
\]

Different thresholds $\tau$ are evaluated to determine the optimal number of features that maximizes classification accuracy. The resulting reduced feature set is then used to train the final classifier.

\subsection{Machine Learning Models}

After feature extraction and selection, we apply supervised machine learning models to classify the MRI volumes into two tumor categories: low-grade glioma (LGG) and high-grade glioma (HGG).

\textbf{Random Forest.}

Random Forest is an ensemble learning method that constructs multiple decision trees and aggregates their predictions. Given a set of training samples

\[
\{(x_i,y_i)\}_{i=1}^{n},
\]

where $x_i \in \mathbb{R}^{d}$ represents the topological feature vector and $y_i \in \{0,1\}$ denotes the class label, the Random Forest classifier predicts the class label using majority voting across $T$ decision trees

\[
\hat{y} = \text{mode}\{T_1(x),T_2(x),\dots,T_T(x)\}.
\]

Random Forest models are robust to noise and can effectively capture nonlinear relationships between features.

\textbf{Extreme Gradient Boosting (XGBoost).}

XGBoost is a gradient boosting algorithm that builds an additive model of decision trees. The prediction for an input feature vector $x_i$ is given by

\[
\hat{y}_i = \sum_{k=1}^{K} f_k(x_i),
\]

where $f_k$ represents the $k$-th decision tree. The model is trained by minimizing the regularized objective function

\[
\mathcal{L} =
\sum_{i=1}^{n} l(y_i,\hat{y}_i)
+
\sum_{k=1}^{K} \Omega(f_k),
\]

where $l$ is the loss function and $\Omega$ is a regularization term that controls model complexity.

Compared to deep learning models that require extensive preprocessing and data augmentation, our approach directly utilizes topological feature vectors derived from the original MRI volumes. Since persistent homology is invariant under transformations such as rotation and reflection, no data augmentation is required. This results in a computationally efficient and interpretable classification framework.

\subsection{Hyperparameters}

The performance of ensemble machine learning models depends on the selection of appropriate hyperparameters. In this study, we use Random Forest and Extreme Gradient Boosting (XGBoost) classifiers to learn discriminative patterns from the extracted topological features. The hyperparameters of both models were empirically selected based on preliminary experiments to achieve stable classification performance while avoiding overfitting.

Random Forest is trained using multiple decision trees constructed from bootstrapped samples of the training data. The final prediction is obtained through majority voting among the individual trees. To increase model robustness, we employ a relatively large number of trees and an entropy-based splitting criterion.

XGBoost is a gradient boosting algorithm that sequentially constructs decision trees to minimize a regularized objective function. Compared to traditional boosting methods, XGBoost includes additional regularization mechanisms and efficient tree construction strategies that improve both accuracy and computational efficiency.

The main hyperparameters used for the Random Forest and XGBoost models are summarized in Table~\ref{tab:hyperparameters}.

\begin{table}[h]
\centering
\caption{Hyperparameters used for Random Forest and XGBoost classifiers.}
\label{tab:hyperparameters}
\begin{tabular}{lll}
\hline
\textbf{Model} & \textbf{Hyperparameter} & \textbf{Value} \\
\hline
\multirow{4}{*}{Random Forest} 
 & Number of trees ($n\_estimators$) & 300 \\
 & Split criterion & Entropy \\
 & Minimum samples for split ($min\_samples\_split$) & 10 \\
 & Random seed ($random\_state$) & 0 \\
\hline
\multirow{7}{*}{XGBoost} 
 & Number of boosting trees ($n\_estimators$) & 1000 \\
 & Maximum tree depth ($max\_depth$) & 25 \\
 & Learning rate ($learning\_rate$) & 0.1 \\
 & Column sampling per tree ($colsample\_bytree$) & 0.4 \\
 & Column sampling per level ($colsample\_bylevel$) & 0.4 \\
 & Evaluation metric ($eval\_metric$) & mlogloss \\
 & Random seed ($random\_state$) & 0 \\
\hline
\end{tabular}
\end{table}

The relatively large number of trees in both models allows the algorithms to capture complex nonlinear relationships among the extracted Betti features. The entropy criterion in Random Forest improves the quality of feature splits, while the depth and learning rate parameters in XGBoost control the complexity of the boosting process. Column subsampling parameters in XGBoost further help reduce overfitting by randomly selecting subsets of features when constructing each tree.

\begin{algorithm}[htbp]
\small
\SetAlgoNlRelativeSize{0}
\DontPrintSemicolon
\caption{Topological Feature Extraction and Classification for BraTS2020 MRI}
\label{alg:tda_brats}

\KwIn{
3D MRI dataset $\mathcal{D}=\{(\chi_i,y_i)\}_{i=1}^{N}$ where $\chi_i \in \mathbb{R}^{155\times240\times240}$,
homology dimensions $k\in\{0,1,2\}$,
number of filtration thresholds $N_f=100$
}

\KwOut{
Trained classifier and evaluation metrics
}

\textbf{Step 1: MRI Volume Preparation} \\
\For{$i \gets 1$ \KwTo $N$}{
Load MRI volume $\chi_i$ with size $155\times240\times240$ \;
Normalize voxel intensities to range $[0,255]$ \;
Assign class label $y_i \in \{\text{LGG},\text{HGG}\}$ \;
}

\textbf{Step 2: Cubical Filtration Construction} \\
\For{$i \gets 1$ \KwTo $N$}{
Construct sublevel filtration
\[
X_1 \subset X_2 \subset \dots \subset X_{N_f}
\]
where
\[
X_n=\{\Delta_{ijk} \subset \chi_i \mid \Upsilon_{ijk}\le t_n\}
\]
}

\textbf{Step 3: Persistent Homology Computation} \\
\For{each MRI volume $\chi_i$}{
\For{$k\in\{0,1,2\}$}{
Compute persistence diagram
\[
PD_k(\chi_i)=\{(b_\sigma,d_\sigma)\}
\]
}
}

\textbf{Step 4: Betti Feature Extraction} \\
\For{$k\in\{0,1,2\}$}{
Compute Betti curves
\[
\vec{\beta}_k(\chi_i)=[\beta_k(t_1),\dots,\beta_k(t_{100})]
\]
}

Construct topological feature vector

\[
\mathbf{f}_i=[\vec{\beta}_0(\chi_i),\vec{\beta}_1(\chi_i),\vec{\beta}_2(\chi_i)]
\in\mathbb{R}^{300}
\]

\textbf{Step 5: Feature Selection} \\

Train a preliminary ensemble model and compute feature importance scores

\[
I_j = \text{importance}(f_j)
\]

Select feature subset

\[
S_\tau = \{f_j \mid I_j \ge \tau\}
\]

\textbf{Step 6: Machine Learning Classification}

\textbf{Random Forest Model}
\[
\hat{y} = \text{mode}\{T_1(x),T_2(x),\dots,T_{300}(x)\}
\]

\textbf{XGBoost Model}

\[
\hat{y}_i=\sum_{k=1}^{K} f_k(x_i)
\]

Train models using selected topological features.

\textbf{Step 7: Model Evaluation}

Perform cross-validation on the dataset.

Compute classification metrics:

\[
\text{Accuracy},\; \text{Precision},\; \text{Recall},\; \text{F1-score},\; \text{AUC}
\]

Generate confusion matrix for model predictions.

\Return Trained models and evaluation metrics

\end{algorithm}

\section{Experiment}

\subsection{Dataset}

The experiments in this study were conducted using the Brain Tumor Segmentation 2020 (BraTS2020) dataset, a widely used benchmark for brain tumor analysis and medical image computing research. The dataset was released as part of the Multimodal Brain Tumor Segmentation Challenge and contains pre-operative magnetic resonance imaging (MRI) scans of patients diagnosed with gliomas.

BraTS2020 consists of 369 subjects collected from multiple institutions and imaging scanners, providing a diverse dataset for robust algorithm evaluation. The dataset includes two clinically relevant tumor grades: High-Grade Glioma (HGG) and Low-Grade Glioma (LGG). Among the available subjects, 293 cases correspond to HGG and 76 cases correspond to LGG. Each subject includes four MRI modalities: T1-weighted (T1), contrast-enhanced T1-weighted (T1ce), T2-weighted (T2), and Fluid Attenuated Inversion Recovery (FLAIR). To ensure consistency across scans, all images were preprocessed by the challenge organizers using skull stripping, co-registration to a common anatomical template, and intensity normalization.

In this study, we utilize the FLAIR modality because it highlights tumor-associated edema and abnormal tissue regions more clearly than other modalities, making it particularly suitable for tumor analysis. Each MRI scan is provided as a three-dimensional volume with spatial dimensions of $240 \times 240 \times 155$ voxels and stored in NIfTI ($.nii$ or $.nii.gz$) format.

To focus on the most informative regions of the brain and reduce computational complexity, we use the central portion of each MRI volume. Specifically, slices indexed from 30 to 90 are extracted from the original 155 slices to construct the working 3D volume. Persistent homology is then applied to these volumes to extract topological descriptors in homology dimensions $k=0,1,$ and $2$, corresponding to connected components, loops or tunnels, and cavities, respectively.

The resulting topological features, represented through Betti curves, are subsequently used to construct machine learning models for binary classification between High-Grade Glioma (HGG) and Low-Grade Glioma (LGG).

\subsection{Experimental Setup}
\noindent \textbf{Training--Test Split:} The dataset is divided into training and testing subsets using an 80:20 ratio, where 80\% of the samples are used to train the machine learning models and the remaining 20\% are reserved for performance evaluation. This split allows the models to learn representative patterns from the training data while ensuring that the final performance is assessed on unseen test samples.
\smallskip

\noindent \textbf{No Data Augmentation:} 
In contrast to conventional CNN-based and deep learning approaches that depend heavily on extensive data augmentation strategies to mitigate limited or imbalanced training data~\cite{goutam2022comprehensive}, the proposed model framework does not require any form of data augmentation. The extracted topological descriptors are inherently invariant to geometric transformations such as rotation and flipping, as well as to minor intensity variations, making the model robust to noise and small perturbations in MRI scans. This eliminates the need for augmentation, significantly reduces computational overhead, and improves training efficiency while maintaining strong generalization performance.


\noindent \textbf{Runtime Efficiency and Platform:} 
All experiments were conducted on a personal laptop equipped with an Apple M1 system-on-chip, featuring an 8-core CPU (4 performance cores and 4 efficiency cores) and 16~GB of unified memory. The implementation was carried out in Python using the TensorFlow/Keras framework, and all models were trained and evaluated on macOS. 

\begin{figure*}[t!]
    \centering
    \subfloat[\scriptsize Radar plot illustrating the performance of the XGBoost classifier using different Betti feature combinations, both with and without feature selection.\label{fig:xgb-radar}]{
        \includegraphics[width=0.45\linewidth]{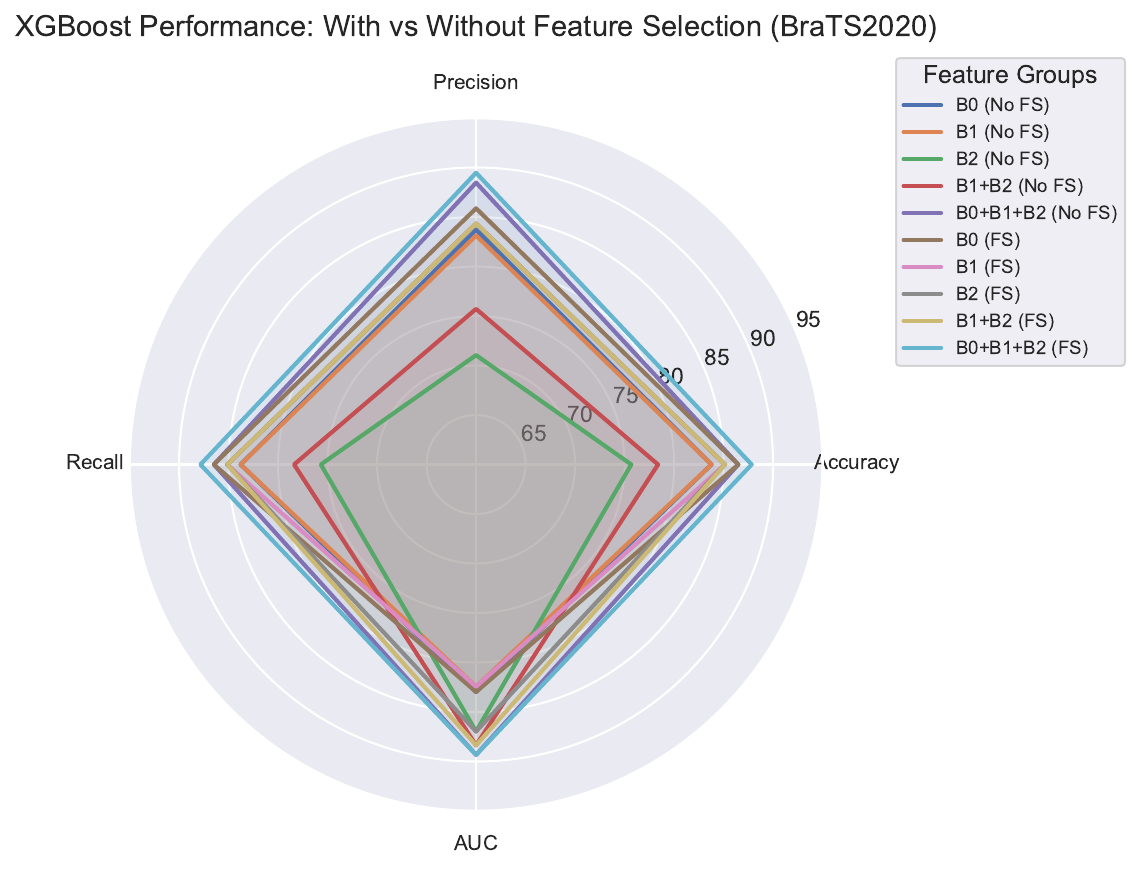}}
    \hfill
    \subfloat[\scriptsize Radar plot illustrating the performance of the Random Forest classifier using different Betti feature combinations, both with and without feature selection.\label{fig:rf-radar}]{
        \includegraphics[width=0.45\linewidth]{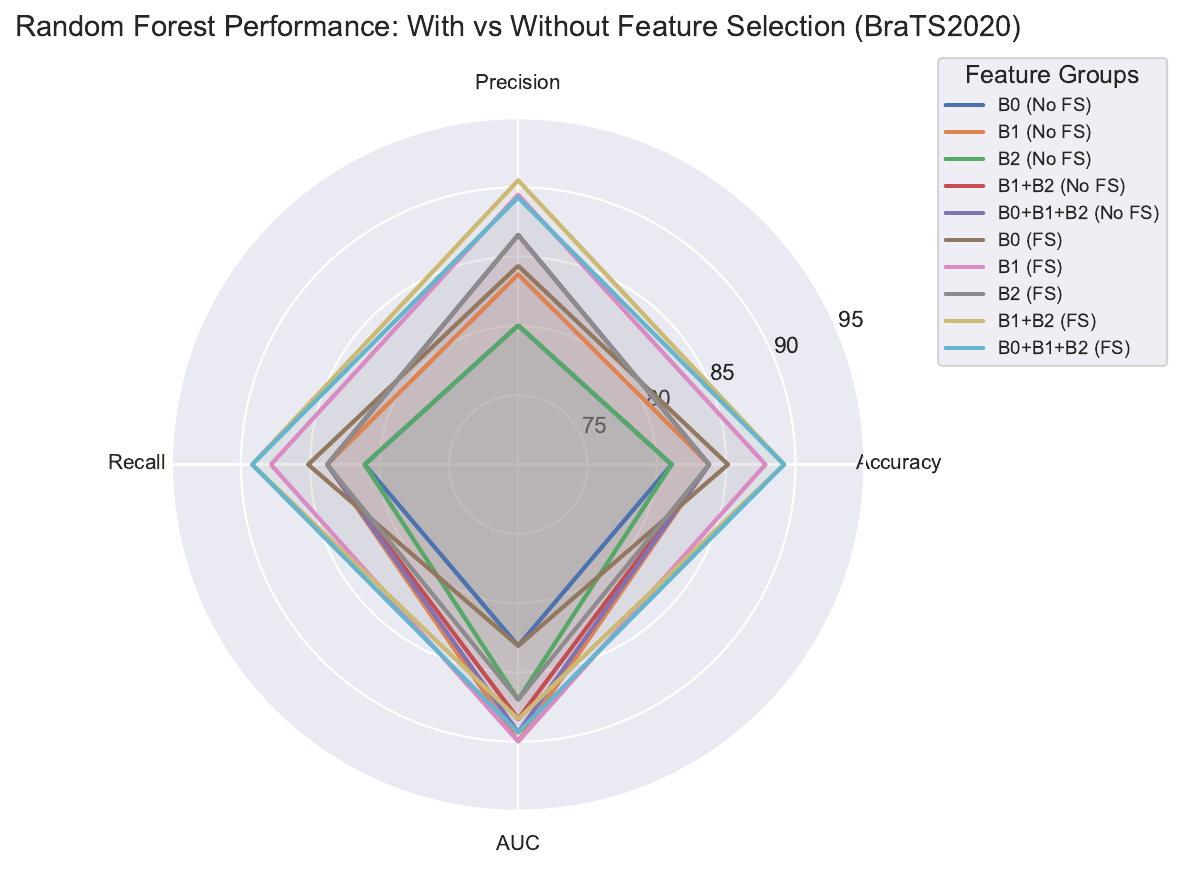}}
    
    \caption{\footnotesize Radar plot comparison of classification performance for XGBoost and Random Forest models using topological features derived from persistent homology on the BraTS2020 3D MRI dataset. Each axis represents one evaluation metric: accuracy, precision, recall, and AUC. The curves correspond to different Betti feature configurations, including Betti-0, Betti-1, Betti-2, and their combinations. Results are shown both with and without feature selection, highlighting the impact of selecting the most informative topological descriptors. The radar plots demonstrate that combining Betti features generally improves model performance, while feature selection further enhances classification accuracy and precision.}
    
    \label{fig:tda-radar-comparison}
\end{figure*}

\section{Results}\label{sec:results}

This section presents the experimental results obtained using topological features extracted from 3D FLAIR MRI volumes of the BraTS 2020 dataset. The proposed framework utilizes persistent homology to derive interpretable Betti-based descriptors, namely Betti-0 (B0), Betti-1 (B1), and Betti-2 (B2), which capture connected components, loops, and void structures in the volumetric data. These features were evaluated using two classical machine learning classifiers: XGBoost and Random Forest. Experiments were conducted both with and without feature selection to assess the impact of dimensionality reduction on classification performance. Visualization of the model performance, both without and with feature selection, is presented in \Cref{fig:tda-radar-comparison}.

\subsection{Performance without Feature Selection}

Table~\ref{tab:xgboost_results} and Table~\ref{tab:rf_results} summarize the classification performance using the complete set of extracted Betti features. When individual Betti descriptors were evaluated independently, both classifiers achieved competitive performance. For XGBoost, the B0 and B1 feature sets achieved an accuracy of 83.78\%, while B2 features resulted in a lower accuracy of 75.68\%. Similarly, the Random Forest classifier achieved an accuracy of 83.78\% using B1 features, demonstrating the discriminative capability of loop-based topological structures for tumor classification.

Combining multiple Betti descriptors further improved the classification performance. In particular, the combination of B0, B1, and B2 features achieved the highest accuracy for XGBoost (86.49\%) with an AUC of 89.33\%. A similar trend was observed for Random Forest, where the combined Betti features achieved an accuracy of 83.78\%. These results indicate that integrating multiple topological descriptors provides a richer representation of tumor morphology in 3D MRI data.

\subsection{Performance with Feature Selection}

To reduce feature redundancy and improve computational efficiency, feature selection was applied to identify the most informative topological descriptors. The results with selected features are also presented in Table~\ref{tab:xgboost_results} and Table~\ref{tab:rf_results}. 

For XGBoost, feature selection improved classification performance across most feature groups. The best performance was achieved using the combined B0+B1+B2 feature set with 42 selected features, resulting in an accuracy of 87.84\% and an AUC of 89.33\%. This demonstrates that a compact subset of topological features can effectively capture the structural characteristics of brain tumors while reducing dimensionality.

The Random Forest classifier showed further improvement after feature selection. The highest classification accuracy of 89.19\% was obtained using both the B1+B2 feature combination (42 selected features) and the combined B0+B1+B2 feature set (14 selected features). In addition, the B1+B2 feature combination achieved the highest precision of 90.50\%, indicating strong discriminative capability when loop and void-based topological features are jointly considered.

\subsection{Discussion}

Overall, the experimental results highlight the effectiveness of topological descriptors derived from persistent homology for brain tumor classification. The results demonstrate that Betti-based features extracted from 3D MRI volumes provide meaningful structural information that can be leveraged by classical machine learning models. Furthermore, feature selection significantly reduces the number of features while improving classification accuracy, which enhances the computational efficiency of the proposed framework.

Among the evaluated models, Random Forest with selected Betti features achieved the best overall performance with an accuracy of 89.19\%. These findings suggest that interpretable topological features can serve as a powerful alternative to high-dimensional deep learning representations, particularly in scenarios where computational efficiency and model interpretability are important considerations.

\begin{table}[ht]
\centering
\caption{Performance of XGBoost using Betti features extracted via persistent homology from 3D FLAIR MRI volumes.}
\label{tab:xgboost_results}
\begin{tabular}{lccccc}
\toprule
\textbf{Feature Set} & \textbf{\# Features} & \textbf{Accuracy (\%)} & \textbf{Precision (\%)} & \textbf{Recall (\%)} & \textbf{AUC (\%)} \\
\midrule
\multicolumn{6}{c}{\textit{Without Feature Selection}} \\
\midrule
B0 & 100 & 83.78 & 83.72 & 83.78 & 82.97 \\
B1 & 100 & 83.78 & 83.13 & 83.78 & 82.44 \\
B2 & 100 & 75.68 & 71.05 & 75.68 & 86.96 \\
B1 + B2 & 200 & 78.38 & 75.72 & 78.38 & 88.36 \\
B0 + B1 + B2 & 300 & 86.49 & 88.47 & 86.49 & 89.33 \\
\midrule
\multicolumn{6}{c}{\textit{With Feature Selection}} \\
\midrule
B0 & 13 & 86.49 & 85.88 & 86.49 & 82.97 \\
B1 & 73 & 85.14 & 84.34 & 85.14 & 82.44 \\
B2 & 6 & 85.14 & 84.34 & 85.14 & 86.96 \\
B1 + B2 & 160 & 85.14 & 84.34 & 85.14 & 88.36 \\
B0 + B1 + B2 & 42 & \textbf{87.84} & \textbf{89.47} & \textbf{87.84} & \textbf{89.33} \\
\bottomrule
\end{tabular}
\end{table}

\begin{table}[ht]
\centering
\caption{Performance of Random Forest using Betti features extracted via persistent homology from 3D FLAIR MRI volumes.}
\label{tab:rf_results}
\begin{tabular}{lccccc}
\toprule
\textbf{Feature Set} & \textbf{\# Features} & \textbf{Accuracy (\%)} & \textbf{Precision (\%)} & \textbf{Recall (\%)} & \textbf{AUC (\%)} \\
\midrule
\multicolumn{6}{c}{\textit{Without Feature Selection}} \\
\midrule
B0 & 100 & 81.08 & 80.04 & 81.08 & 83.08 \\
B1 & 100 & 83.78 & 83.72 & 83.78 & 89.98 \\
B2 & 100 & 81.08 & 80.04 & 81.08 & 86.96 \\
B1 + B2 & 200 & 83.78 & 86.56 & 83.78 & 88.36 \\
B0 + B1 + B2 & 300 & 83.78 & 86.56 & 83.78 & 89.33 \\
\midrule
\multicolumn{6}{c}{\textit{With Feature Selection}} \\
\midrule
B0 & 8 & 85.14 & 84.34 & 85.15 & 83.08 \\
B1 & 26 & 87.84 & 89.47 & 87.84 & 89.98 \\
B2 & 5 & 83.78 & 86.56 & 83.78 & 86.96 \\
B1 + B2 & 42 & \textbf{89.19} & \textbf{90.50} & \textbf{89.19} & 88.36 \\
B0 + B1 + B2 & 14 & \textbf{89.19} & 89.27 & \textbf{89.19} & \textbf{89.33} \\
\bottomrule
\end{tabular}
\end{table}

\section{Conclusion}\label{sec:conclusion}

In this study, we presented a topology-driven framework for brain tumor classification using persistent homology applied to three-dimensional (3D) MRI volumes from the BraTS 2020 dataset. By leveraging Topological Data Analysis (TDA), we extracted interpretable Betti-based descriptors, including Betti-0, Betti-1, and Betti-2 features, which capture fundamental structural properties such as connected components, loops, and voids within the volumetric images. These features provide a compact representation of complex tumor morphology while preserving important topological information.

The extracted topological features were used to train classical machine learning classifiers, including XGBoost and Random Forest, for binary classification of high-grade glioma (HGG) and low-grade glioma (LGG). Experimental results demonstrated that combining Betti features improves classification performance, and feature selection further enhances accuracy while significantly reducing the number of features. The best performance was achieved using the Random Forest classifier with selected Betti features, reaching an accuracy of 89.19\% and strong precision and AUC values.

Compared with conventional deep learning approaches, the proposed method offers several advantages, including improved interpretability, lower computational complexity, and reduced reliance on large-scale training data or data augmentation strategies. By directly analyzing the topology of 3D MRI volumes, the proposed framework effectively captures global structural characteristics of brain tumors that are difficult to represent using traditional intensity-based features.

Future work will explore integrating topological descriptors with deep learning representations to further enhance classification performance. Additionally, extending the proposed framework to multi-modal MRI data and other medical imaging applications may provide deeper insights into complex anatomical structures and disease patterns.

 \section*{Declarations}

 \textbf{Funding} \\
 The author received no financial support for the research, authorship, or publication of this work.

 \vspace{2mm}
 \textbf{Author's Contribution} \\
 Faisal Ahmed conceptualized the study, downloaded the data, prepared the code, performed the data analysis and wrote the manuscript. Faisal Ahmed reviewed and approved the final version of the manuscript. 

  \vspace{2mm}
 \textbf{Acknowledgement} \\
The authors utilized an online platform to check and correct grammatical errors and to improve sentence readability.

 \vspace{2mm}
 \textbf{Conflict of interest/Competing interests} \\
 The authors declare no conflict of interest.

 \vspace{2mm}
 \textbf{Ethics approval and consent to participate} \\
 Not applicable. This study did not involve human participants or animals, and publicly available datasets were used.

 \vspace{2mm}
 \textbf{Consent for publication} \\
 Not applicable.

 \vspace{2mm}
 \textbf{Data availability} \\
 The datasets used in this study are publicly available online. 

 \vspace{2mm}
 \textbf{Materials availability} \\
 Not applicable.


\newpage


\bibliography{refs}
\end{document}